\documentclass{article}
\usepackage{spconf,amsmath,graphicx}
\usepackage{graphicx}

\title{FACE ANTI-SPOOFING BASED ON COLOR TEXTURE ANALYSIS}
\vspace{-6mm}
\name{Zinelabidine Boulkenafet, Jukka Komulainen, Abdenour Hadid}
\vspace{-6mm}
\address{Center for Machine Vision Research, University of Oulu, Finland}
\begin{document}
%\ninept
%
\maketitle
\vspace{-6mm}
\begin{abstract}
Research on face spoofing detection has mainly been focused on analyzing the luminance of the face images, hence discarding the chrominance information which can be useful for discriminating fake faces from genuine ones. In this work, we propose a new face anti-spoofing method based on color texture analysis. We analyze the joint color-texture information from the luminance and the chrominance channels using a color local binary pattern descriptor. More specifically, the feature histograms are extracted from each image band separately. Extensive experiments on two benchmark datasets, namely CASIA face anti-spoofing and Replay-Attack databases, showed excellent results compared to the state-of-the-art. Most importantly, our inter-database evaluation depicts that the proposed approach showed very promising generalization capabilities.
\end{abstract}
%
%\begin{keywords}
%Face anti-spoofing, Color texture analysis, Local Binary Patterns

%\end{keywords}
%
\vspace{-2mm}
\section{Introduction}
\vspace{-2mm}
\label{sec:intro}

Nowadays, it is not a secret that most of existing face recognition systems are vulnerable  to spoofing attacks.  A spoofing attack occurs when someone tries to bypass a face biometric system by presenting a fake face in front of the camera. For instance, in \cite{spoof_OSN}, researchers  inspected the threat of the online social networks based facial disclosure against  the  latest version of  six commercial face authentication systems (Face Unlock, Facelock Pro, Visidon, Veriface, Luxand Blink and FastAccess). While on average only 39\% of the images published on social networks could be successfully used for spoofing, the relatively small number of vulnerable images was enough to fool face authentication software of 77\% of the 74 users. Also, in a live demonstration during the International Conference on Biometric (ICB 2013), a female intruder with a specific make-up succeeded in fooling a face recognition system~\cite{tabula_rasa}. These two examples among others highlight the vulnerability of face recognition systems to spoofing attacks. 
\begin{figure}[t]
	\centering
		\includegraphics[height=160pt,width=250pt]{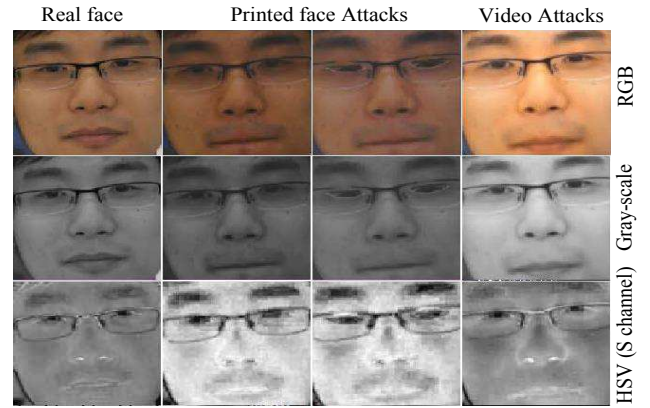}
	\label{exemple_of_face_images}
	\vspace{-8mm}
	\caption{RGB, gray-scale and HSV color space  of real and fake face images}
\end{figure}
Research on face spoofing detection has mainly been focusing on analyzing gray-scale images \cite{micro_texture} \cite{replay} \cite{lbp_casia} and hence discarding the color information which can be useful for discriminating fake faces from genuine ones. While the proposed methods have shown promising results on individual datasets \cite{replay} \cite{lbp_casia}, the generalization capabilities of these methods have been questionable \cite{motion_casia}. Facial texture analysis from gray-scale images might provide sufficient means to reveal the recapturing artifacts of fake faces if the image resolution (quality) is good enough to capture the fine details of the observed face. However, if we take a look at the cropped facial images of a genuine human face and corresponding fake ones in Figure \ref{exemple_of_face_images}, it is basically impossible to explicitly name any textural differences between them because the input image resolution is not high enough. Human eye is indeed more sensitive to luminance than to chroma, thus fake faces still look very similar to the genuine ones when the same facial images are shown in color (see Figure \ref{exemple_of_face_images}). However, if only the corresponding chroma component is considered, some characteristic differences can be already noticed.

Inspired by the aforementioned observations, we propose, in this work, a new face anti-spoofing method based on color texture analysis. The color Local Binary Patterns (LBP) descriptor proposed in \cite{color_lbp} is used to extract the joint color-texture information from the face images. In this descriptor, the uniform LBP histograms are extracted from the individual image bands. Subsequently, these histograms are concatenated to form the final descriptor. To gain insight into which color space is more discriminative to distinguish real face from fake ones, we considered three color spaces, namely $RGB$, $HSV$ and $YC_bC_r$. Extensive experiments on two challenging benchmark databases, namely CASIA face anti-spoofing and Replay-Attack databases, clearly indicate that color texture based method outperforms gray-scale counterparts in detecting various types of spoofing attacks. Moreover, our inter-database experiments showed that the proposed approach yields in very promising generalization capabilities compared to state-of-the-art methods.

%The rest of this paper is organized as follows: in Section \ref{sec:lbp} we describe our proposed face anti-spoofing, in Section \ref{sec:databases},  we present the spoofing databases used in our experiments. Section \ref{sec:experiments} discuss our experiments and the obtained results, while Section \ref{sec:conclusion} conclude the paper.   

%%%%%%%%%%%%%%%%%%%%%%%%%%%%%%%%%%%%%%%%%%%%%%%%%%%%%%%%%%%%%%%%%%%%%%%%%%%%%%%%%%%%%%%%%%%%%%%%%%%%%%%%%%%%%%%%%%%%%%%%%%%%%%%%%%%%%%%%%%%%%%%%%%%%%%%%%%%%%%%%%%%%%%%%%%
	\vspace{-4mm}
\section{COLOR LBP BASED FACE ANTI-SPOOFING }
	\vspace{-2mm}
\label{sec:lbp}

Face spoofing attacks are most likely performed by displaying the targeted face using prints or video screens. Attack attempts with low facial texture quality (e.g. mobile phone) can be detected by analyzing the texture and the quality of the gray-scale images. However, as shown in Figure \ref{exemple_of_face_images}, it is reasonable to assume that fake faces of higher quality are harder or nearly impossible to detect using only luminance information of webcam-quality images.

Fortunately, the color reproduction (gamut) of different display media, e.g. photographs, video displays and masks, is limited compared to genuine faces. Thus, the presented fake faces suffer from spoofing medium dependent color. Furthermore, a recaptured face image is likely to contain local variations of color due to other imperfections in the reproduction process of the targeted face. Both the display medium dependent color gamut signatures and the local chroma variations (noise) can be described by analyzing the color texture of the chroma channels. Since the chrominance channels are separated from the luminance information, they are also more tolerant to illumination variation assuming that the acquisition conditions are reasonable. We aim to explore how the aforementioned visual cues can be used for face anti-spoofing. We investigate which color models provide the most useful micro-texture representations by extracting LBP descriptions from the different color spaces.  

\begin{figure*}[t]
	\begin{center}
		\includegraphics[height=135pt]{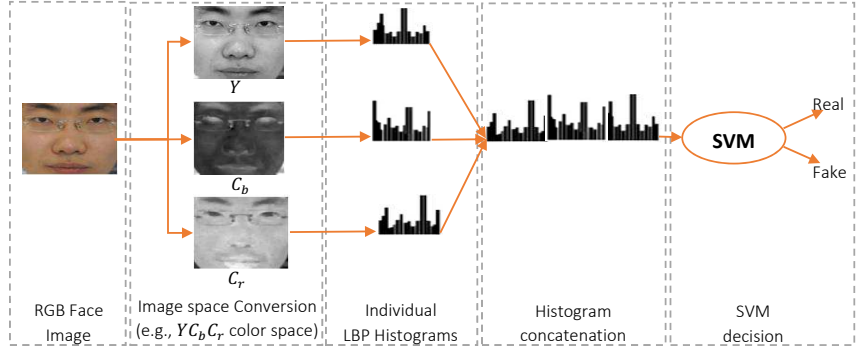}		
	\end{center}
		\vspace{-6mm}
	\caption{Architecture of the proposed face anti-spoofing approach}
	\label{architecture}
\end{figure*}

	\vspace{-4mm}
\subsection{Color Spaces}
%$RGB$ is the most used color space for representing color images. It is, however, not a perceptually uniform space in the sense that differences between colors, measured by the Euclidean distance, in the three dimensional RGB space do not correspond to color differences as perceived by humans \cite{human_color_perception}. For this reason other uniform or approximately uniform color spaces has been defined. Most of these color spaces are based on the separation between the luminance and the chrominance information. 
$RGB$ is the most used color space for sensing, representation and displaying of color images. However,  its application in image analysis is quite limited due to the high correlation between the three color components (red, green and blue) and the imperfect separation of the luminance and chrominance information.

In this work, we considered two other color spaces to explore the color texture information in addition to RGB: the $HSV$ and the $YC_bC_r$. Both of these color spaces are based on the separation of the luminance and the chrominance information. In the $HSV$ color space, the hue and the saturation dimensions define the chrominance of the image while the value dimension corresponds to the luminance. The $YC_bC_r$ space separates the RGB components into luminance (Y), chrominance blue ($C_b$) and chrominance red ($C_r$). More details about these color spaces can be found e.g. in  \cite{color_image}. 
	\vspace{-3mm}
\subsection{Texture Representation}  
The LBP descriptor proposed by Ojala et al. \cite{lbp_ojala} is a highly discriminative gray-scale texture descriptor. For each pixel in an image, a binary code is computed by thresholding a circularly symmetric neighborhood with the value of the central pixel. Finally, a histogram is created to collect the occurrences of different binary patterns. LBP was originally intended to handle gray-scale images but was later extended to exploit also color information. In \cite{color_lbp}, a simple yet efficient color LBP descriptor was proposed. The LBP operator is applied on each color band. The obtained histograms are concatenated to form the final color descriptor as depicted in Figure \ref{architecture}. The LBP pattern of a pixel $(x,y)$ extracted from the image band (i) can be written as follows:   
\begin{equation}
LBP_{P,R}^{(i)}(x,y)=
\begin{cases}
\sum_{n=0}^{P-1} \delta(r_{n}^{(i)}-r_{c}^{(i)})\times 2^n ~\text{if}~ U^{(i)}<=2
\\
P(P-1)+2~ otherwise
\end{cases}
\end{equation}

where
\begin{equation}
\begin{split}
U^{(i)}=&\left|\delta(r_{p-1}^{(i)}-r_{c}^{(i)})-\delta(r_{0}^{(i)}-r_{c}^{(i)})\right|+\\
&\sum_{n=1}^{P}\left|\delta(r_{n}^{(i)}-r_{c}^{(i)})-\delta(r_{n-1}^{(i)}-r_{c}^{(i)})\right|
\end{split}
\end{equation}
and $\delta(x)= 1$ if $x>=0$, otherwise $0$. $r_c$ and $r_n(n=0,...,P-1)$ denote the intensity values of the central pixel $(x,y)$ and its $P$ neighborhood pixels located at the circle of radius $R$ ($R>0$).  

To summarize, let $I$ be a face image represented in a color space $S$ ($S \in\{RGB, HSV, YC_bC_r\}$) and, let $H_s^{(i)}, \{i=1:M\}$ be its uniform LBP histogram extracted from the $M$ channel of the space $S$. The color LBP features of the image $I$ represented in the space $S$ can be defined by:
\begin{equation}
\label{color_LBP} 
H_s=[H_s^{(1)}...H_s^{(M)}]
\end{equation} 
To detect spoofing  attacks, the color LBP features extracted from  face images are fed into a Support Vector Machine (SVM) classifier. The general  diagram of our proposed approach for detecting spoofing attacks is depicted in Figure \ref{architecture}.
%%%%%%%%%%%%%%%%%%%%%%%%%%%%%%%%%%%%%%%%%%%%%%%%%%%%%%%%%%%%%%%%%%%%%%%%%%%%%%%%%%%%%%%%%%%%%%%%%%%%%%%%%%%%%%%%%%%%%%%%%%%%%%%%%%%%%%%%%%%%%%%%%%%%%%%%%%%%%%%%%%%%%%%%%%%
	\vspace{-6mm}
\section{EXPERIMENTAL ANALYSIS}
	\vspace{-2mm}
\label{sec:experiments} 
In this section, we first describe briefly the two benchmark datasets and the setup that was used in our experiments. Then, an in-depth analysis of the color texture based approach is provided.
\vspace{-2mm}
\subsection{ Experimental Setup}
To assess the effectiveness of our proposed anti-spoofing technique, we considered the CASIA Face Anti-Spofing and the Replay-Attack databases. These two datasets are the most challenging  face anti-spoofing benchmark databases that consist of recordings of real client accesses and various spoofing attack attempts.

The CASIA Face Anti-spoofing (CASIA-FA) database \cite{casia} contains video recordings of real and fake faces. The real faces were recorded from 50 genuine subjects, where the fake faces were made from high quality records of genuine faces. Three fake face attacks were designed: \textit{warped photo attacks}, \textit{cut photo attacks}, and \textit{video attacks}. Both of real face access and fake face attacks were recorded using three camera resolutions: \textit{low resolution}, \textit{normal resolution}, and \textit{high resolution}. The 50 subjects were divided into two subsets for training and testing (20 and 30, respectively). The test protocol consists of seven scenarios: The first three scenarios were designed to study the effect of the imaging quality: low  quality (1), normal quality (2) and high quality (3). The next three scenarios correspond to warped photo attacks (4), cut photo attacks (5) and video attacks (6). Finally, the overall scenario (7) was designed to give an overall evaluation performance by combining all the aforementioned scenarios.

The Replay-Attack database \cite{replay} consists of video recordings of  real-access and attack attempts to 50 clients. Each person in the database was recorded a number of videos in two illumination conditions: \textit{controlled} and \textit{adverse}. Under the same conditions, a high resolution pictures and videos were taken for each person. Three type of attacks were designed: (1) \textit{print attacks}, (2)~\textit{mobile attacks}, and (3) \textit{highdef attacks}. According to the support used in presenting the fake face devices in front of the camera, the attacks were divided into: \textit{hand based attacks} (the attack devices were held by the operator) and \textit{fixed-support attacks} (the attack devices were set on a fixed support). For evaluation, the total set of videos is divided into three non-overlapping subsets for training, development and testing.  

In our experiments, we followed the defined protocols of the two databases which allows a fair comparison against the state-of-the-art. On CASIA-FA database, the model parameters are trained and tuned using four-fold subject-disjoint cross-validation on the training set and the results are reported in terms of Equal Error Rate (EER) on the test set. Replay-Attack database provides also a separate validation set for tuning the model parameters. Thus, the results are given in terms of EER on the development set and the Half Total Error Rate (HTER) on the test set.

In all our experiments, we used the $LBP_{8,1}^{(i)}$ operator (i.e., $P=8$ and $R=1$) to extract the textural features from the normalized ($64 \times 64$)  face images.
To capture both of the appearance and the motion variation of the face images, we average the features within a time windows of three seconds and four seconds on CASIA-FA and Replay-Attack databases, respectively. In order to get more training data, these time windows are taken with  two seconds overlap in the training stage. In the test stage, only the average of the features within the first time window is used to classify each video. The classification was done using a Support Vector Machine \cite{libsvm} (SVM) with RBF Kernel. 
		\vspace{-1mm}
\subsection{Results}
	\vspace{-1mm}
\label{sec:experiments}
 Table \ref{perf_casia} and Table \ref{perf_replay} present the results of different LBP based color texture descriptions and their gray-scale counterparts. From these results, we can clearly see that the color texture features significantly improve the performance compared to the gray-scale LBP-based countermeasure. When comparing the different color spaces, $YC_bC_r$ based representation yields to the best overall performance. The color LBP features extracted from the $YC_bC_r$ space improves the performance on CASIA-FA and Replay-Attack databases by 64.5 \% and 81.4 \%, respectively, compared to the gray-scale LBP features.
			
From Table \ref{perf_casia}, we can also observe that the features extracted from the $HSV$ color space seem to be more effective against video attacks than those extracted from the $YC_bC_r$ color space. Thus,  we studied the benefits of combining the two color texture representations by fusing them at feature level. The color LBP descriptions from the two color spaces were concatenated, thus the size of the resulting histogram is $59 \times 3 \times 2$. The results in Table \ref{perf_casia} and Table \ref{perf_replay} indicate that a significant performance enhancement is obtained, thus confirming the benefits of combining the different facial color texture representations.

\vspace{-4mm}
\begin{table}[h]
\small
		\centering
	\caption{The performance in therm of EER \% of the gray-scale LBP and the color LBP descriptors on CASIA FA database}
	 \label{perf_casia}
		\setlength{\tabcolsep}{0.14cm}
		\begin{tabular}{|c|c|c|c|c|c|c|c|}
			\hline
			Method & 1 & 2& 3& 4 & 5 & 6& 7\\
			\hline \hline
			Gray-scale-LBP& 16.5& 17.2 & 23.0& 24.7 &16.7& 27.0 &24.8\\
					\hline 
			RGB-LBP& 17.9 & 17.6 & 11.1 & 18.0& 12.5 & 14.3&16.1\\
			\hline 
			HSV-LBP& 12.2 & 10.3& 14.4 & 11.0 &11.1  & \textbf{7.8}&13.4\\
			\hline 
			YCbCr-LBP& 9.0 & \textbf{9.5 }&\textbf{6.0} & 9.9&8.8  & 12.1 &8.8 \\
			\hline
			\hline
			YCbCr+HSV-LBP& \textbf{7.8}& 10.1 &6.4 &\textbf{ 7.5} &\textbf{5.4}  & 8.1 &\textbf{6.2}\\
			\hline 			
	 \end{tabular}	
	\end{table} 	
	\vspace{-6mm}
		\begin{table}[h]
		\small
			\centering
	\caption{The performance of the gray-scale LBP and the color LBP descriptors on Replay-Attack database}
		%\small
	\label{perf_replay}
	\setlength{\tabcolsep}{0.7cm}
		\begin{tabular}{|c|c|c|}
			\hline
	    Method & EER \%& HTER\%\\
			\hline \hline
			Gray-scale-LBP&15.3 &15.6 \\
					\hline 
			RGB-LBP& 5.0 & 6.6 \\
			\hline 
			HSV-LBP&6.4  & 7.0 \\
			\hline 
			YCbCr-LBP&  0.7 & 3.3 \\
			\hline 
			\hline
			YCbCr+HSV-LBP&  \textbf{0.4} & \textbf{2.9} \\
			\hline
	 \end{tabular}	
	\end{table}  
	
Table \ref{comp} compares the performance of our proposed countermeasure against the state-of-the-art face anti-spoofing methods. From this table, we can notice that our method outperforms the state-of-the-art results on the challenging CASIA-FA database, and yields in very competitive results on the Replay-Attack database.
	
\vspace{-4mm}
		\begin{table}[h]
		\small
			\centering
	\caption{Comparison between the proposed countermeasure and state-of-the-art methods}
	%	\small
	\label{comp}
	\setlength{\tabcolsep}{0.35cm}
		\begin{tabular}{|c|c|c|c|}
		\hline
	       & \multicolumn{2}{c|}{Replay-Attack} & CASIA\\
			\cline{2-4}
	    Method & EER \% & HTER\%& EER\%\\
			\hline \hline
			IQA based \cite{IQA}&- &- & 32.4\\
			\hline
			CDD \cite{low_level_high_level}& - & - & 11.8\\
			\hline
			DOG (baseline)\cite{casia}& - & -&17.0\\
			\hline						
			Motion+LBP \cite{motion_lbp}& 4.5 & 5.1&- \\
			\hline 
			Motion \cite{motion_casia}&11.6 &11.7&26.6 \\
					\hline 
			LBP \cite{replay}& 13.9 & 13.8&18.2 \\
			\hline 
			LBP-TOP \cite{lbp_top_replay}& 7.8   & 7.6&10.6 \\
			\hline 
			Motion Mag \cite{motion_mag}& \textbf{0.2}& \textbf{0.0}&14.4 \\
			\hline 
			%Deep  Learning\cite{Deep_learning}& 6.1& 2.1&7.3 \\
			\hline 
			Proposed method&  0.4 & 2.9&\textbf{6.2}\\
			\hline
	 \end{tabular}	
	\end{table} 
\vspace{-3mm}
		\begin{table}[h]
			\centering
	\caption{Inter-Test results in term of  HTER \%  on CASIA and Replay-Attack databases}
		\small
	\label{cross_database}
	\setlength{\tabcolsep}{0.22cm}
		\begin{tabular}{|c|c|c|c|c|}
		\hline
		 & \multicolumn{2}{c|}{Replay-Attack}&\multicolumn{2}{c|}{CASIA }\\
		\cline{2-5}
		Method& $Dev$&$Test$&$Train$&$Test$\\
		\hline
     Motion\cite{motion_casia}&50.2&50.2&47.7&48.2\\
    \hline
    LBP \cite{motion_casia}&44.9&47.0&57.3&57.9\\
    \hline
    LBP-TOP \cite{motion_casia}&48.9&50.6&60.0&61.3\\
    \hline
		Motion-Mag\cite{motion_mag}&50.0&50.20&43.8&50.3\\
		 \hline
		%Deep-learning\cite{Deep_learning}&48.2&48.8&45.7&45.4\\
		\hline
		Our method (SVM-RBF)&22.5 &20.6 &47.5&43.9\\
		    \hline
			Our method (SVM-linear)&\textbf{17.7}&\textbf{16.7}&\textbf{38.6}&\textbf{37.6}\\
		    \hline
	 \end{tabular}	
	\end{table} 	
	To gain insight into the generalization capabilities of our proposed method, we conducted a cross-database evaluation. In these experiments, the countermeasure was trained and tuned with one database (CASIA-FA or Replay-Attack) and then tested on the other database. The results of these experiments are summarized in Table \ref{cross_database}. 

In the first experiment, we evaluated the performance on  CASIA-FA database while training and tuning the countermeasure on the Replay-Attack database. Table \ref{cross_database} reports an HTER values of  $47.5 \%$ and $43.9 \%$  on the training and the testing sets, respectively. While, in the second experiment, when the  countermeasure is trained and tuned on CASIA-FA database and then tested on Replay-Attack database, the HTER values on the development and the test sets are  $22.5\%$ and $20.6\%$, respectively. Although these results are very competitive to  those of state-of-the-art methods, specially on Replay-Attack database, they are still degraded compared to the intra-test results (when the countermeasure is trained and tested on the same database).  

Complex classifiers, like SVM-RBF, might be more sensitive to over-fitting than simpler classification schemes. The two face anti-spoofing benchmark datasets are rather small and the variations in the provided data are also limited which increases the chance of over-fitting with powerful texture features and complex classification schemes. Inspired by the observations e.g. in \cite{replay} \cite{motion_lbp}, we proposed to mitigate this problem by using linear SVM instead of SVM-RBF

The experiments using linear SVM models show  very interesting  results compared to  those of the SVM-RBF models.  On the CASIA-FA database, the HTER values  on the training  and the testing  sets have been reduced to $38.6\%$ and $37.6\%$, respectively.  On Replay-Attack database, the HTER values have been reduced to $17.7\%$ and $16.7\%$ (on the development and the test sets, respectively), that are comparable to  those obtained with the gray-scale LBP descriptor in the intra-test evaluation ($15.3 \%$ and $15.6\%$). 

The model optimized on the Replay-Attack dataset is not able to generalize as well as the model based on the CASIA FA. The reason behind this is that the CASIA FA dataset contains more variations in the collected data (e.g. imaging quality and proximity between the camera and face) compared to the Replay-Attack database. Therefore, the model optimized for Replay-Attack database has difficulty to perform well in the new environmental conditions. One way to deal with this problem is to train countermeasure with a joint training set by combining the train set of both databases, as described in \cite{motion_casia}. 

	\vspace{-2mm}
\section{CONCLUSION}
	\vspace{-2mm}
\label{sec:conclusion}
In this paper, we proposed to approach the problem of face anti-spoofing from the color texture analysis point of view. We investigated which of color spaces, $RGB$, $HSV$ and $YC_bC_r$, provide useful face representations for describing the color texture differences between genuine faces and fake ones. The effectiveness of the different color texture representations was studied by extracting color LBP features from the individual image channels. Extensive experiments on two challenging spoofing databases, CASIA-FA and Replay-Attack, showed excellent results. On CASIA-FA database, the face representation based on the combination of $HSV$ and $YC_bC_r$ color spaces beat the state-of-the-art. Furthermore, in our inter-database evaluation, the proposed approach showed very promising generalization capabilities. As future work, more experiments should be conducted in order to get more insight into the color texture based face anti-spoofing and to derive problem-specific facial color representations.

\section*{Acknowledgments}
The financial support of the Academy of Finland is fully acknowledged. 

\bibliographystyle{IEEEbib}
\bibliography{refs}

\begin{thebibliography}{10}

\bibitem{spoof_OSN}
Yan Li, Ke~Xu, Qiang Yan, Yingjiu Li, and Robert~H. Deng,
\newblock ``Understanding osn-based facial disclosure against face
  authentication systems,''
\newblock in {\em Proceedings of the 9th ACM Symposium on Information, Computer
  and Communications Security}, New York, NY, USA, 2014, ASIA CCS '14, pp.
  413--424, ACM.

\bibitem{tabula_rasa}
Tabula Rasa,
\newblock ``Tabula rasa spoofing challenge,''
\newblock Tech. {R}ep., 2013,
\newblock
  http://www.tabularasa-euproject.org/evaluations/tabula-rasa-spoofingchallenge-2013.

\bibitem{micro_texture}
J.~Maatta, A.~Hadid, and M.~Pietikainen,
\newblock ``Face spoofing detection from single images using micro-texture
  analysis,''
\newblock in {\em International Joint Conference on Biometrics (IJCB)}, Oct
  2011, pp. 1--7.

\bibitem{replay}
I.~Chingovska, A.~Anjos, and S.~Marcel,
\newblock ``On the effectiveness of local binary patterns in face
  anti-spoofing,''
\newblock in {\em International Conference of the Biometrics Special Interest
  Group (BIOSIG)}, Sept 2012, pp. 1--7.

\bibitem{lbp_casia}
Tiago Freitas~Pereira, Jukka Komulainen, Andre Anjos, Jose De~Martino, Abdenour
  Hadid, Matti Pietikainen, and Sebastien Marcel,
\newblock ``Face liveness detection using dynamic texture,''
\newblock {\em EURASIP Journal on Image and Video Processing}, vol. 2014, no.
  1, pp. 2, 2014.

\bibitem{motion_casia}
T.~de~Freitas~Pereira, A.~Anjos, J.M. De~Martino, and S.~Marcel,
\newblock ``Can face anti-spoofing countermeasures work in a real world
  scenario?,''
\newblock in {\em International Conference on Biometrics (ICB)}, June 2013, pp.
  1--8.

\bibitem{color_lbp}
Jae~Young Choi, K.N. Plataniotis, and Yong~Man Ro,
\newblock ``Using colour local binary pattern features for face recognition,''
\newblock in {\em IEEE International Conference on Image Processing (ICIP)},
  Sept 2010, pp. 4541--4544.

\bibitem{color_image}
Konstantinos N.~Plataniotis Rastislav~Lukac,
\newblock {\em Color Image Processing: Methods and Applications}, vol.~8,
\newblock New York CRC, 2007.

\bibitem{lbp_ojala}
T.~Ojala, M.~Pietikainen, and T.~Maenpaa,
\newblock ``Multiresolution gray-scale and rotation invariant texture
  classification with local binary patterns,''
\newblock {\em IEEE Transactions on Pattern Analysis and Machine Intelligence
  (PAMI)}, vol. 24, no. 7, pp. 971--987, Jul 2002.

\bibitem{casia}
Zhiwei Zhang, Junjie Yan, Sifei Liu, Zhen Lei, Dong Yi, and S.Z. Li,
\newblock ``A face antispoofing database with diverse attacks,''
\newblock in {\em International Conference on Biometrics (ICB)}, March 2012,
  pp. 26--31.

\bibitem{libsvm}
Chih-Chung Chang and Chih-Jen Lin,
\newblock ``{LIBSVM}: A library for support vector machines,''
\newblock {\em ACM Transactions on Intelligent Systems and Technology}, vol. 2,
  pp. 27:1--27:27, 2011.

\bibitem{IQA}
Javier Galbally and Sebastien Marcel,
\newblock ``Face anti-spoofing based on general image quality assessment,''
\newblock in {\em International Conference on Pattern Recognition (ICPR)}, Aug
  2014, pp. 1173--1178.

\bibitem{low_level_high_level}
Jianwei Yang, Zhen Lei, Shengcai Liao, and S.Z. Li,
\newblock ``Face liveness detection with component dependent descriptor,''
\newblock in {\em Biometrics (ICB), 2013 International Conference on}, June
  2013, pp. 1--6.

\bibitem{motion_lbp}
J.~Komulainen, A.~Hadid, M.~Pietikainen, A.~Anjos, and S.~Marcel,
\newblock ``Complementary countermeasures for detecting scenic face spoofing
  attacks,''
\newblock in {\em International Conference on Biometrics (ICB)}, June 2013, pp.
  1--7.

\bibitem{lbp_top_replay}
Tiago de~Freitas~Pereira, AndrÃ© Anjos, JosÃ©Mario De~Martino, and
  SÃ©bastien Marcel,
\newblock ``Lbp top based countermeasure against face spoofing attacks,''
\newblock in {\em Computer Vision ACCV 2012 Workshops}, Jong-Il Park and Junmo
  Kim, Eds., vol. 7728 of {\em Lecture Notes in Computer Science}, pp.
  121--132. Springer Berlin Heidelberg, 2013.

\bibitem{motion_mag}
Bharadwaj Samarth, Dhamecha~Tejas I, Vatsa Mayank, and Singh Richa,
\newblock ``Face anti-spooﬁng via motion magniﬁcation and multifeature
  videolet aggregation,''
\newblock Tech. {R}ep., University of Delhi, Department of Computer Science and
  Engineering, 5 2014,
\newblock https://repository.iiitd.edu.in/jspui/handle/123456789/138.

\end{thebibliography}
\end{document}